\documentclass[10pt,twocolumn,letterpaper]{article}

\usepackage[pagenumbers]{iccv} 

\usepackage{booktabs} 

\usepackage{graphicx}
\usepackage{amsmath}

\usepackage[utf8]{inputenc} 
\usepackage[T1]{fontenc}    
\usepackage{url}            
\usepackage{amsfonts}       
\usepackage{nicefrac}       
\usepackage{microtype}      
\usepackage{xcolor}         
\usepackage{bm}         
\usepackage{amsmath}         
\usepackage{graphicx}

\usepackage{multirow}
\usepackage{makecell}
\usepackage{pifont}
\usepackage{wrapfig}

\usepackage{caption}
\usepackage{comment}
\usepackage{makecell} 

\usepackage{xspace}

\makeatletter
\DeclareRobustCommand\onedot{\futurelet\@let@token\@onedot}
\def\@onedot{\ifx\@let@token.\else.\null\fi\xspace}

\def\eg{\emph{e.g}\onedot} 
\def\ie{\emph{i.e}\onedot}

\makeatother

\definecolor{iccvblue}{rgb}{0.21,0.49,0.74}
\usepackage[pagebackref,breaklinks,colorlinks,allcolors=iccvblue]{hyperref}

\newcommand{\nickname}{HumanGif}

\title{HumanGif: Single-View Human Diffusion with Generative Prior}

\author{
Shoukang Hu\textsuperscript{\rm 1} \quad
Takuya Narihira\textsuperscript{\rm 1} \quad
Kazumi Fukuda\textsuperscript{\rm 1} \quad
Ryosuke Sawata\textsuperscript{\rm 1} \quad \\
Takashi Shibuya\textsuperscript{\rm 1} \quad
Yuki Mitsufuji\textsuperscript{\rm 1,2}\\
\textsuperscript{\rm 1}Sony AI \quad
\textsuperscript{\rm 2}Sony Group Corporation \\
}

\begin{document}

\twocolumn[{
    \renewcommand\twocolumn[1][]{#1}%
    \maketitle
    \vspace{-20pt}
    \begin{center}
        \centering
        \includegraphics[width=1.0\textwidth]{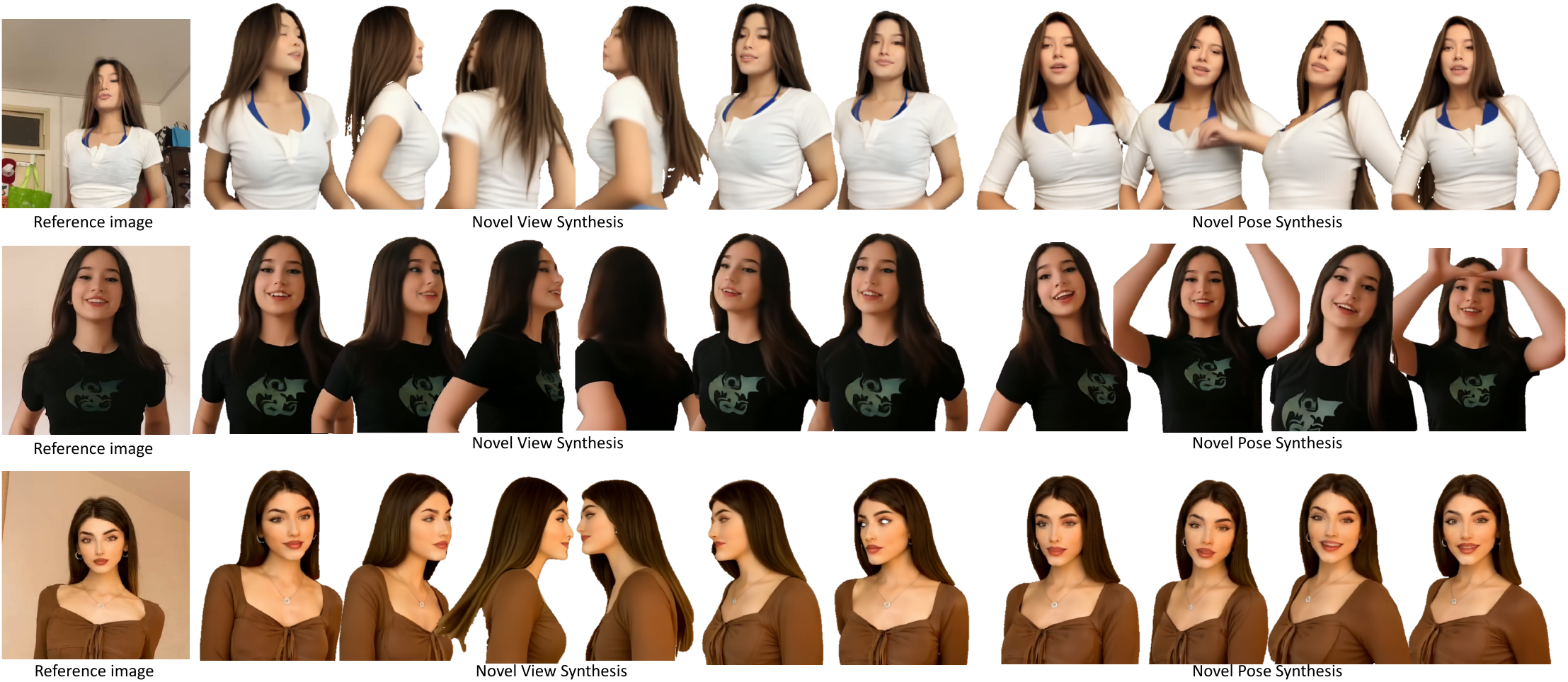}
        \captionof{figure}{\nickname{} is a single-view human diffusion model. By inheriting generative prior, \nickname{} synthesizes realistic view and pose-consistent images.}
        \label{fig:teaser}
    \end{center}
}]

\maketitle
\begin{abstract}
Previous 3D human creation methods have made significant progress in synthesizing view-consistent and temporally aligned results from sparse-view images or monocular videos.
However, it remains challenging to produce perpetually realistic, view-consistent, and temporally coherent human avatars from a single image, as limited information is available in the single-view input setting. 
Motivated by the success of 2D character animation, we propose \textbf{\nickname}, a single-view human diffusion model with generative prior.
Specifically, we formulate the single-view-based 3D human novel view and pose synthesis as a single-view-conditioned human diffusion process, utilizing generative priors from foundational diffusion models to complement the missing information. 
To ensure fine-grained and consistent novel view and pose synthesis, we introduce a Human NeRF module in \nickname{} to learn spatially aligned features from the input image, implicitly capturing the relative camera and human pose transformation.
Furthermore, we introduce an image-level loss during optimization to bridge the gap between latent and image spaces in diffusion models.
Extensive experiments on RenderPeople, DNA-Rendering, THuman 2.1, and TikTok datasets demonstrate that \nickname\ achieves the best perceptual performance, with better generalizability for novel view and pose synthesis.
 Our code is available at \url{https://github.com/skhu101/HumanGif}. 
\end{abstract}

\section{Introduction}

Synthesizing 3D human performers with consistent novel views and poses holds extensive utility across various domains, including AR/VR, video games, and movie production.
Recent methods enable the novel view and pose synthesis of 3D human avatars from sparse-view human videos~\cite{neuralbody, chen2021animatable, peng2022animatable, xu2021h, noguchi2021neural, weng_humannerf_2022_cvpr, su2021nerf, jiang2022selfrecon, jiang2022neuman, wang2022arah, hu2024gauhuman, chen2023primdiffusion, hong2022eva3d, hu2023humanliff, hong2022avatarclip, li2024synthesizing}, 
with Neural Radiance Field~\cite{mildenhall2021nerf} or Gaussian Splatting-based~\cite{kerbl20233d} representation. 
While impressive novel view and pose synthesis results are achieved from sparse-view videos, generating perpetually realistic, view-consistent, and temporally coherent human avatars from a single image~\cite{hu2023sherf, dong2023ivs, weng2023zeroavatar, liao2023high, huang2023tech, pan2024humansplat, wang2024geneman, ho2024sith, chen2024generalizable, cha2023generating, casas2023smplitex} remains challenging as limited information is available.

To address the challenge of complementing information missing from a single image, a line of research~\cite{weng2023zeroavatar, huang2023tech, pan2024humansplat, casas2023smplitex} focuses on the novel view synthesis from a single human image by complementing human images or UV textures at unseen views (\eg, back view) through foundational models (\eg, T2I diffusion models).
To further learn view-consistent and temporally aligned human avatars from a single image, another line of research~\cite{hu2023sherf, shao2024360} proposes synthesizing novel views and poses from a single image by learning a generalizable HumanNeRF or a human diffusion model from scratch.
However, such frameworks normally require multi-view human image collections as training datasets, and their performance is closely tied to the quality and scale of these datasets.
Although efforts to create high-quality multi-view human image datasets~\cite{tao2021function4d, han2023high, cheng2023dna, li2021ai, cai2022humman, peng2021neural, ho2023learning, tang2023human, alldieck2018video, ionescu2013human3, mono-3dhp2017, peng2022animatable, habermann2020deepcap, habermann2021real, huang2024wildavatar, icsik2023humanrf, yang2023synbody} have been accelerated in recent years, the number of human subjects in these datasets remains significantly smaller in comparison with the size of multi-view object and scene image datasets~\cite{deitke2023objaverse,wu2023omniobject3d,yu2023mvimgnet,tung2025megascenes, dai2017scannet, zhou2018stereo}. 
This data disparity limits the generalization performance of single-image-based 3D human modeling frameworks.

To mitigate the data-sparsity dilemma, recent research~\cite{bhunia2023person, karras2023dreampose,wang2023disco, chang2023magicpose, xu2024magicanimate,hu2024animate,zhu2024champ,zhang2024mimicmotion,kim2024tcan, liyuan2024cfsynthesis} in 2D character animation involves generative prior (inherit pre-trained weights) from Text-to-image (T2I) diffusion models (\eg, Stable Diffusion~\cite{rombach2021highresolution}) to assist the single-view based 2D human animation (novel pose synthesis).
By inheriting the generative priors from diffusion models, these approaches demonstrate impressive generalization performance for novel human subjects and poses in 2D character animation tasks (e.g., dancing video generation) using only a modest amount of training data, such as hundreds of monocular dancing videos~\cite{jafarian2021learning}.
Motivated by these successes, it is desirable to explore the use of generative prior from T2I diffusion models in the single-view-based 3D human novel view/pose synthesis task.

In this work, we propose \textbf{\nickname}, a single-view conditioned human diffusion with generative prior. 
Specifically, we formulate the single-view-based 3D human novel view and pose synthesis as a single-view-conditioned human diffusion process, utilizing generative priors from foundational diffusion models to complement the missing information. 
Yet, this promising avenue comes with two primary barriers: 
1) How to spatially and temporally align the learned human avatar with the reference image and target human poses?
2) How to ensure the performance achieved in the latent diffusion space is equally effective in the image-level space?
In particular, one potential solution to the first challenge is to inject cross-attention modules into diffusion models to learn spatial and temporal transformation from the reference image and target pose images.
However, our experiments reveal that human diffusion models built for 2D character animation struggle with a) learning detailed information (e.g., logos on T-shirts) even when such details are present in the reference image, and b) achieving view-consistent and temporally aligned results.

To address the first challenge, we introduce a more spatially and temporally aligned conditioning signal drawing inspiration from 3D human reconstruction methods~\cite{weng_humannerf_2022_cvpr, hu2023sherf, wu2024reconfusion}. 
Specifically, we learn a Human NeRF module to transform the human subject from the reference pose space to the target pose space using a parametric human SMPL model~\cite{SMPL:2015}.
The rendered human images in the target space provide explicit conditional information, reducing the difficulty of transforming information from the reference image space to the target space.
In addition, we encode camera pose information into a Plücker ray representation~\cite{sitzmann2021light}, serving as a conditional signal of relative camera poses.
To further enhance the spatial and temporal consistency, we incorporate a class embedding to utilize attention modules for novel view and pose synthesis tasks.
For the second challenge, we observe a discrepancy between the latent and image space, attributed to the Variational Autoencoder (VAE)~\cite{kingma2019introduction} used in diffusion models.
Inspired by the Radiance field rendering loss employed in 3D object diffusion models~\cite{muller2023diffrf}, we construct an image-level loss by mapping the noise latent to the image space through a VAE decoder.
This ensures consistent optimizations in both latent and image spaces.
With the incorporated modules, our \nickname{} successfully recovers fine-grained information from the reference image and achieves the best perceptually realistic, view-consistent, and temporally coherent results.
Our main contributions are as follows:
\begin{enumerate}
    \item We introduce \nickname{}, a single-view-conditioned human diffusion model that incorporates generative priors to compensate for information missing from the single input image.
    \item To learn perpetually realistic, view-consistent, and temporally aligned 3D human avatars, we render human images in the target space, along with Plücker ray representation to serve as a more spatially and temporally aligned conditioning signal. 
    \item To bridge the gap between latent and image spaces, we propose an image-level loss, which decodes the diffused latent space into image space during training, ensuring consistent optimization across both domains.
    \item Extensive experiments on human datasets demonstrate that \nickname{} outperforms baseline methods in perceptual quality of novel view and pose synthesis.
\end{enumerate}

\section{Related Work}

\noindent \textbf{Diffusion Models.}
Diffusion models~\cite{sohl2015deep, ho2020denoising} have demonstrated remarkable performance in image synthesis tasks, especially for text-to-image (T2I) generation~\cite{balaji2022ediff, nichol2021glide, saharia2022photorealistic, ramesh2022hierarchical, huang2023composer, rombach2022high}.
To improve sample quality for conditional generation, classifier guidance~\cite{dhariwal2021diffusion} and classifier-free guidance~\cite{ho2022classifier} are introduced by leveraging an explicitly trained classifier or score estimates from both conditional and unconditional generation. 
In this work, we propose utilizing the generative prior from latent diffusion models (e.g., Stable Diffusion~\cite{rombach2022high}) for novel view and pose synthesis of 3D humans from a single image.

\noindent \textbf{Novel View/Pose Synthesis from a Single Human Image.}
Although 2D human modelling~\cite{albahar2021pose, lewis2021tryongan, sarkar2021humangan, men2020controllable, fu2022stylegan, jiang2022text2human, jiang2023text2performer, fu2023unitedhuman} has made substantial progress, it is still challenging to synthesizing 3D humans from monocular inputs, especially for a single human image input~\cite{saito2019pifu, saito2020pifuhd, he2020geo, li2020robust, dong2022pina, li2020monocular, bozic2021neural, yang2021s3, bhatnagar2020combining, bhatnagar2020loopreg, huang2020arch, he2021arch++, zheng2021pamir, xiu2022icon, xiu2022econ, alldieck2022photorealistic, corona2022structured}.
To complement the information missing from a single image input, these approaches~\cite{weng2023zeroavatar, huang2023tech, pan2024humansplat, wang2024geneman, chen2024generalizable, choi2022mononhr, zhang2024sifu, ho2024sith, dong2023ivs, cha2023generating, casas2023smplitex} typically leverage text-to-image (T2I) diffusion models, such as Stable Diffusion~\cite{rombach2021highresolution}, by employing subject-specific Score Distillation Sampling (SDS)~\cite{poole2022dreamfusion} or fine-tuning a T2I diffusion model conditioned on a front-view image using multi-view human datasets or 3D scan datasets. 
This enables the synthesis of human images from unseen viewpoints, such as generating a back view from a given front view.
To further learn view-consistent and temporally aligned human avatars from a single image, recent research~\cite{hu2023sherf, shao2024360} leverages the strong modeling capabilities of deep neural networks to learn generalizable features for Neural Radiance Field (NeRF) or diffusion models, enabling the synthesis of novel views and poses in a feed-forward manner. 
However, the scale of multi-view human datasets limits their generalization ability.
In this work, we propose to utilize the generative prior from diffusion models to produce perpetually realistic, view-consistent, and temporally coherent human avatars from a single image.

\noindent \textbf{Diffusion Models for 2D Character Animation.}
2D Character Animation aims at generating temporally coherent animation videos from one or more static human images~\cite{albahar2023single, cao2024dreamavatar, chan2019everybody, fu2022stylegan, jiang2023humangen, prokudin2021smplpix, ren2020deep, sarkar2020neural, Siarohin_2019_NeurIPS, siarohin2021motion, yoon2021pose, yu2023bidirectionally, zhang2022exploring, zhao2022thin}.
With advancements in text-to-image diffusion models, recent studies~\cite{bhunia2023person, karras2023dreampose, wang2023disco, chang2023magicpose, xu2024magicanimate, hu2024animate, zhu2024champ, zhang2024mimicmotion, kim2024tcan, liyuan2024cfsynthesis} have investigated the use of diffusion models for 2D character animation.
These approaches normally use a reference human image and a sequence of target pose images (\eg, OpenPose~\cite{8765346}, DWPose~\cite{yang2023effective}, DensePose~\cite{guler2018densepose}, or SMPL pose~\cite{SMPL:2015}) as conditional inputs to pose encoders or ControlNet~\cite{zhang2023adding}, generating a sequence of target images aligned with the reference human image and target pose images.
Motivated by the success of diffusion-based methods in 2D character animation, we adapt these methodologies for 3D/4D human modeling from a single image. 
Human4DiT~\cite{shao2024360} introduced a diffusion model that generates multi-view human animation videos from a single image, utilizing factorized image, view, and temporal modules. 
However, the code for this model is not publicly available. 
We identified that our baseline multi-view video diffusion model could not capture fine-grained details and maintain multi-view consistency. 
We show that integrating a generative prior and our proposed modules into the diffusion pipeline effectively resolves these challenges.
A concurrent work GAS~\cite{lu2025gas} also utilizes diffusion models to enhance the novel view and pose synthesis performance from a Human NeRF module.

\section{Our Approach}
\begin{figure*}[t]
    \centering
    \vspace{-6mm}
    \includegraphics[width=16cm]{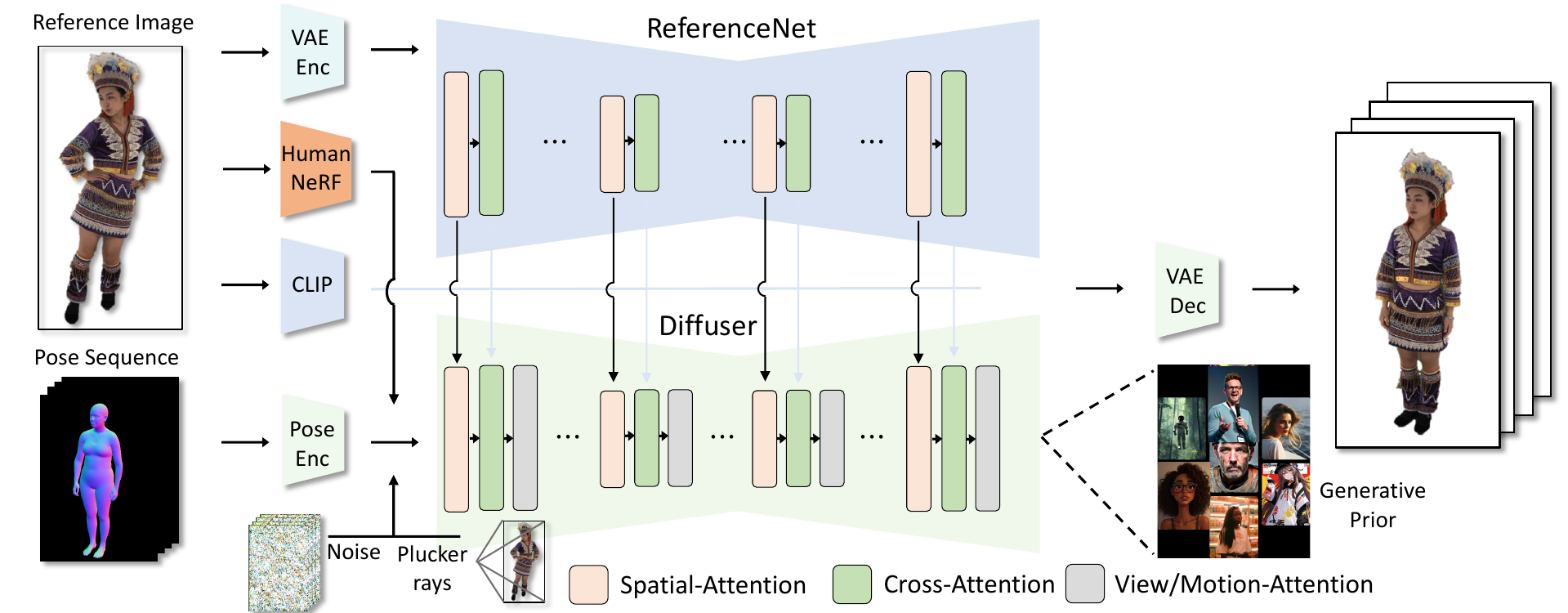}
    \setlength{\abovecaptionskip}{0cm}
    \caption{\textbf{\nickname{} Framework}. Given a single input human image and a target pose sequence, our \nickname{} produces a sequence of target images aligned with the input image and target human poses. To synthesize view-consistent and temporally coherent outputs, our \nickname{} proposes incorporating a generative prior, a Human NeRF module, Plücker ray representation, image-level loss, view/temporal attention.} 
\label{fig: overview}
\vspace{-2mm}
\end{figure*}

Our proposed \nickname, as illustrated in Fig.~\ref{fig: overview}, learns a single-view conditioned human diffusion model for novel view and pose synthesis by utilizing a generative prior from Stable Diffusion.
Specifically, given a reference human image and a target human pose sequence (estimated from a human video or generated from other modalities like text, or audio), \nickname{} aims to predict a sequence of human pose images aligned with the person in the reference image and motions observed in the target human pose sequence. 

\subsection{Preliminary}
\label{sec:preliminary}

\noindent \textbf{Latent Diffusion Model (LDM)}~\cite{rombach2022high} presents a novel class of diffusion models by integrating two distinct stochastic processes, \ie, diffusion and denoising, directly within the latent space.
LDM learns a variational autoencoder (VAE)~\cite{kingma2013auto, van2017neural} to establish the mapping from image space to latent space, reducing the diffusion model's complexity.
Specifically, give an image $\bm{x}$, the encoder maps the image to a latent representation $\bm{z}_0=\mathcal{E}(\bm{x})$ and the decoder reconstructs it to image space $\bm{x}=\mathcal{D}(\bm{z}_0)$. 
The diffusion process progressively adds Gaussian noise to the data $\bm{z}_0$ following a variance schedule $1-\alpha_{0}, \dots, 1-\alpha_{T}$ specified for different time steps, \ie,
\begin{equation}
\label{eqn:diffusion_process}
    \begin{aligned}
        \bm{z}_{t}=\sqrt{\alpha_t}\bm{z}_{t-1} + \sqrt{1-\alpha_t}\bm{\epsilon}.
    \end{aligned}
\end{equation}
With a sufficiently large number of steps $T$, the diffusion process converges to $\bm{z}_T\sim \mathcal{N}(\bm{0}, \bm{I})$. 

The denoising process learns to denoise $\bm{z}_{t}$ to $\bm{z}_{t-1}$ by predicting the noise $\bm{\epsilon}_{\Phi}(\bm{z}_{t}, t, \bm{c})$ for each denoising step, where $\bm{\epsilon}_{\Phi}(\bm{z}_{t}, t, \bm{c})$ is the output from a neural network (\eg, UNet~\cite{ronneberger2015u}) and $\bm{c}$ denotes conditional signal, \eg, the text embedding from CLIP~\cite{radford2021learning} model.
The diffusion loss $\mathcal{L}_\text{diff}$ is constructed by calculating the expected mean squared error (MSE) over the actual noise $\bm{\epsilon}$ and predicted noise $\bm{\epsilon}_{\Phi}(\bm{z}_{t}, t, \bm{c})$, \ie, 
\begin{equation}
\label{eqn:denoising_process}
    \begin{aligned}
        \mathcal{L}_\text{diff-latent}=\mathbb{E}_{\bm{z}_{t}, \bm{\epsilon}, t, \bm{c}}[w_t\Arrowvert\bm{\epsilon}_{\Phi}(\bm{z}_{t}, t, \bm{c}) - \bm{\epsilon}\Arrowvert^{2}],
    \end{aligned}
\end{equation}
where $w_t$ is the weighting of the loss at time step $t$.

\noindent \textbf{SMPL}~\cite{SMPL:2015} is a parametric human model, $M(\bm{\beta}, \bm{\theta})$, where $\bm{\beta}, \bm{\theta}$ control body shape and pose, respectively. 
In this work, we utilize the Linear Blend Skinning (LBS) algorithm employed in SMPL to transform points from canonical to posed spaces.
For instance, a 3D point $\bm{p}^c$ in the canonical space is transformed into the posed space defined by pose $\bm{\theta}$ as $\bm{p}^\text{tgt} = \sum_{k=1}^{K}w_{k}(\bm{G}_k(\bm{J}, \bm{\theta})\bm{p}^c+\bm{b}_k(\bm{J}, \bm{\theta}, \bm{\beta}))$, where $\bm{J}$ represents $K$ joint locations, $\bm{G}_k(\bm{J}, \bm{\theta})$ is the transformation matrix of joint $k$, $\bm{b}_k(\bm{J}, \bm{\theta}, \bm{\beta})$ is the translation vector of joint $k$, and $w_{k}$ is the linear blend weight.

\noindent \textbf{NeRF}~\cite{mildenhall2021nerf} learns an implicit, continuous function that maps the 3D location $\bm{p}$ and unit direction $\bm{d}$ of a point to its volume density $\bm{\sigma} \in [0, \infty)$ and color value $\bm{c}\in [0,1]^3$, i.e., $F_{\Phi}: (\gamma(\bm{p}), \gamma(\bm{d})) \to (\bm{c}, \bm{\sigma})$, where $F_{\Phi}$ is parameterized by a multi-layer perceptron (MLP) network, $\gamma$ denotes a predefined positional embedding applied to $\bm{p}$ and $\bm{d}$.
To render the RGB color of pixels in the target view, rays are cast from the camera origin $\bm{o}$ through the pixel along the unit direction $\bm{d}$. 
Based on the classical volume rendering~\cite{kajiya1984ray}, the expected color $\hat{C}(\bm{r})$ for a camera ray $\bm{r}(t) = \bm{o} + t\bm{d}$ is computed as 
\begin{equation}
\label{eqn:volume_rendering}
\begin{aligned}
\hat{C}(\bm{r})=\int_{t_n}^{t_f} T(t) \sigma(\bm{r}(t)) \bm{c}(\bm{r}(t), \bm{d})dt,
\end{aligned}
\end{equation}
where $T(t)=\exp(-\int_{t_n}^{t}\sigma(\bm{r}(s))ds)$ denotes the accumulated transmittance along direction $\bm{d}$ from near bound $t_n$ to current position $t$, and $t_f$ represents the far bound.
In practice, the integral is approximated using the quadrature rule~\cite{max1995optical}, which reduces to traditional alpha compositing.

\subsection{Single-view Human Diffusion Model}
\label{sec:architecture}
Motivated by the success of leveraging generative priors from T2I diffusion models in the 2D character animation~\cite{hu2024animate}, we propose utilizing generative priors for single-view-based 3D human novel view/pose synthesis tasks. 
Specifically, we reformulate the single-view-based 3D human novel view and pose synthesis as a single-view-conditioned human diffusion process, utilizing generative priors from foundational diffusion models to complement the missing information.

\noindent \textbf{Denoising UNet.} 
Our backbone, illustrated in Fig.~\ref{fig: overview}, is a denoising U-Net that inherits both the architecture and pre-trained weights (generative prior) from Stable Diffusion (SD) 1.5~\cite{rombach2022high}.
The vanilla SD UNet in Stable Diffusion comprises three main components: downsampling, middle, and upsampling blocks.
Each block integrates multiple interleaved layers, including convolution layers for feature extraction, self-attention layers for spatial feature aggregation, and cross-attention layers that interact with CLIP text embeddings to guide the denoising process.
To process multiple noise latents to produce human images that align with the subject in the given reference image and the specified target poses, we further incorporate the reference image and the target pose sequence through the following modules.

\noindent \textbf{ReferenceNet.}
Building on recent advances in 2D character animation~\cite{hu2024animate}, as illustrated in Fig.~\ref{fig: overview}, we integrate a copy of the SD denoising U-Net as the ReferenceNet to extract features from the reference image.
Specifically, we replace the self-attention layer with a spatial-attention layer, enabling self-attention on the concatenated features from the denoising U-Net and ReferenceNet. 
Additionally, we incorporate a CLIP encoder to extract semantic features, which are fused with the features from the denoising U-Net using cross-attention modules.

\noindent \textbf{Pose Encoder.}
There are various options for defining target poses, including OpenPose, DensePose, DWPose, and SMPL parametric poses.
We follow~\cite{zhu2024champ} to estimate SMPL mesh for the reference image, animate the SMPL mesh with the target SMPL pose parameters, and render normal images as the target guidance signal. 
The SMPL normal pose images are encoded through a Pose Encoder~\cite{hu2024animate}, which contains four convolution layers. 
The Pose Encoder output is added to the noise latent to provide view and pose guidance.

\begin{table*}[h]
\vspace{-5mm}
\setlength{\abovecaptionskip}{0cm}
\caption{Quantitative comparison of our \nickname{} and baseline methods on the RendePeople, DNA-Rendering, THuman2.1 and TikTok datasets. $*$ denotes that checkpoints released from the original work are used for performance evaluation.
}
\centering
\label{tab: main_result}
\begin{tabular}{l|cccc|cccc|c}
\toprule
\multirow{3}*{Method} & \multicolumn{9}{c}{RenderPeople} \\
\cline{2-10}
~ & \multicolumn{4}{c|}{Novel View} & \multicolumn{4}{c|}{Novel Pose} \\
~ & L1$\downarrow$ & PSNR$\uparrow$ & SSIM$\uparrow$ & LPIPS$\downarrow$ & L1$\downarrow$ & PSNR$\uparrow$ & SSIM$\uparrow$ & LPIPS$\downarrow$ & - \\
\midrule
MagicAnimate$^*$~\cite{xu2024magicanimate} & 1.27E-04 & 17.161 & 0.910 & 0.148 & 1.06E-04 & 18.440 & 0.922 & 0.129 & - \\
Champ$^*$~\cite{zhu2024champ} & 7.46E-05 & 16.311 & 0.457 & 0.452 & 1.32E-05 & 23.285 & 0.940 & 0.047 & - \\
\midrule
SHERF~\cite{hu2023sherf} & \underline{9.75E-06} & \textbf{26.128} & \underline{0.934} & \underline{0.063} & \textbf{7.48E-06} & \textbf{27.435} & \underline{0.946} & 0.048 & - \\
Animate Anyone~\cite{hu2024animate} & 2.32E-05 & 21.022 & 0.929 & 0.064 & 1.35E-05 & 24.306 & \underline{0.946} & 0.041 & - \\
Champ~\cite{zhu2024champ} & 2.42E-05 & 21.326 & 0.930 & 0.064 & 1.34E-05 & 25.381 & \textbf{0.952} & \underline{0.037} & - \\
\textbf{\nickname \:(Ours)} & \textbf{9.47E-06} & \underline{25.110} & \textbf{0.951} & \textbf{0.037} & \underline{8.98E-06} & \underline{25.440} & \textbf{0.952} & \textbf{0.034} & - \\
\bottomrule
\multirow{3}*{Method} & \multicolumn{9}{c}{DNA-Rendering} \\
\cline{2-10}
~ & \multicolumn{4}{c|}{Novel View} & \multicolumn{4}{c|}{Novel Pose} \\
~ & L1$\downarrow$ & PSNR$\uparrow$ & SSIM$\uparrow$ & LPIPS$\downarrow$ & L1$\downarrow$ & PSNR$\uparrow$ & SSIM$\uparrow$ & LPIPS$\downarrow$ &  FVD$\downarrow$ \\
\midrule
MagicAnimate$^*$~\cite{xu2024magicanimate} & 3.89E-04 & 7.463 & 0.677 & 0.573 & 3.75E-04 & 7.797 & 0.630 & 0.613 & 100.12 \\
Champ$^*$~\cite{zhu2024champ} & 2.95E-05 & 16.855 & 0.544 & 0.414 & 2.55E-05 & 18.093 & 0.537 & 0.411 & 50.26 \\
\midrule
SHERF~\cite{hu2023sherf} & \textbf{5.73E-06} & \textbf{24.885} & \underline{0.931} & 0.063 & \textbf{5.13E-06} & \textbf{25.560} & \underline{0.914} & \underline{0.050} & 12.01 \\
Animate Anyone~\cite{hu2024animate} & 6.20E-06 & 23.499 & 0.902 & 0.056 & 6.84E-06 & 23.043 & 0.907 & 0.061 & 16.22\\
Champ~\cite{zhu2024champ} & 6.37E-06 & \underline{23.715} & 0.869 & \underline{0.054} & 6.96E-06 & 23.253 & 0.879 & 0.058 & 16.89 \\
\textbf{\nickname \:(Ours)} & \underline{5.82E-06} & 23.686 & \textbf{0.935} & \textbf{0.047} & \underline{5.67E-06} & \underline{24.275} & \textbf{0.935} & \textbf{0.045} & \textbf{9.88} \\
\bottomrule
\multirow{3}*{Method} & \multicolumn{9}{c}{THuman2.1 \& TikTok} \\
\cline{2-10}
~ & \multicolumn{4}{c|}{Novel View} & \multicolumn{4}{c|}{Novel Pose} \\
~ & L1$\downarrow$ & PSNR$\uparrow$ & SSIM$\uparrow$ & LPIPS$\downarrow$ & L1$\downarrow$ & PSNR$\uparrow$ & SSIM$\uparrow$ & LPIPS$\downarrow$ &  FVD$\downarrow$ \\
\midrule
Champ$^*$~\cite{zhu2024champ} & 3.36E-04 & 6.858 & 0.660 & 0.444 & 1.23E-04 & 13.395 & 0.727 & 0.275 & 35.31 \\
\midrule
SHERF~\cite{hu2023sherf} & \underline{1.07E-05} & \underline{25.148} & 0.935 & 0.071 & \textbf{7.63E-05} & \textbf{16.930} & 0.715 & 0.257 & 20.46 \\
Animate Anyone~\cite{hu2024animate} & 1.61E-05 & 21.487 & \underline{0.932} & 0.054 & 9.25E-05 & 14.207 & 0.760 & 0.238 & 42.89 \\
Champ~\cite{zhu2024champ} & 1.13E-05 & 23.828 & 0.945 & \underline{0.040} & \underline{8.60E-05} & 14.761 & \underline{0.768} & \underline{0.231} & 32.93 \\
\textbf{\nickname \:(Ours)} & \textbf{9.71E-06} & \textbf{25.966} & \textbf{0.956} & \textbf{0.029} & 8.77E-05 & \underline{15.044} & \textbf{0.772} & \textbf{0.223} & \textbf{18.39} \\
\bottomrule
\end{tabular}
\end{table*}

\noindent \textbf{Human NeRF.}
We observe two challenges by adopting human diffusion models from 2D character animation tasks: (a) difficulty in capturing fine-grained details (e.g., logos on T-shirts), even when they are present in the reference image, and (b) limited capability to maintain consistency across multiple views and poses.
One underlying reason is that the reference image is encoded into a latent representation and processed by ReferenceNet to extract features, resulting in information loss during this process.
Inspired by prior research in 3D human reconstruction methods~\cite{peng2021neural, hu2023sherf, chen2021animatable, animatablenerf, xu2021h, noguchi2021neural, weng2022humannerf, su2021nerf, jiang2022selfrecon, jiang2022neuman, wang2022arah, kwon2021neural, gao2022mps, choi2022mononhr, zhao2021humannerf, huang2022elicit}, we introduce a more spatially and temporally aligned conditioning signal by rendering human images in the target space.

The input to the Human NeRF module is a single human image $\mathbf{I}^\text{ref}$ along with its corresponding camera parameters $\bm{P}^\text{ref}$ and SMPL pose parameter $\bm{\theta}^\text{ref}$ and shape parameter $\bm{\beta}^\text{ref}$.
The module outputs the rendered human feature image in the target camera view $\bm{P}^\text{tgt}$, corresponding to the target SMPL pose $\bm{\theta}^{tgt}$ and shape $\bm{\beta}^\text{tgt}$.
Specifically, in the target space, we cast rays passing through the camera origin and image pixels, and sample points $\bm{x}^\text{tgt}$ along the cast rays. 
These points $\bm{x}^\text{tgt}$ are transformed into the canonical space $\bm{x}^{c}$ using inverse Linear Blend Skinning (LBS).
Subsequently, hierarchical 3D-aware features are queried from their respective feature extraction modules.
The queried features are concatenated and passed to the NeRF decoder to predict the density $\sigma$ and feature $\bm{c}$ for each sampled point. 
The final pixel features are rendered in the target space through volume rendering, integrating the density and feature values of the sampled 3D points along the rays in the target space.
The details of extracting 3D-aware features are described in the appendix.

\noindent \textbf{Plücker Ray Representation.}
Camera pose information is important for novel view and pose synthesis, as relative camera poses are attributed to the reference-to-target and target-to-target transformation.
We encode camera pose information into a Plücker ray
representation~\cite{sitzmann2021light} and add it to the input noise signal, serving as a conditional signal of camera poses.

\noindent \textbf{View/Temporal Module.}
To learn view-consistent and temporally coherent 3D human avatars, we further integrate a view/temporal attention layer after the spatial-attention and cross-attention components within the Res-Trans block of Denosing UNet. 
Instead of using two separate attention modules for synthesizing views and poses, we adopt a unified attention module but different class embeddings for the two tasks.
Inspired by the efficient temporal attention layer adopted in~\cite{hu2024animate, guo2023animatediff}, we utilize the same architecture for our view/temporal attention layer.
Furthermore, inspired by~\cite{he2025magicman}, we extend the original self-attention in the Denoising UNet to be 3D attention. 
The 3D attention learns view/temporal information by conducting 3D attention in a selected number of feature maps from nearby views or time steps. 

\subsection{Training Detail}
\label{sec:training_detail}

Our training objective $\mathcal{L}$ comprises three components: 1) $\mathcal{L}_\text{diff-latent}$, as shown in Eqn.~\ref{eqn:diffusion_process}, which aligns the learned latent space with the data distribution, 2) $\mathcal{L}_\text{diff-img}$, which focuses on enhancing the quality of the decoded image from the latent space, and 3) $\mathcal{L}_\text{NeRF}$, which regularizes the learned image features using images and human masks from the target space, \ie, 
\begin{equation}
\label{loss: overall}
\begin{aligned}
\mathcal{L} = \mathcal{L}_\text{diff-latent} + \lambda_{1}\mathcal{L}_\text{diff-img} + \lambda_{2}\mathcal{L}_\text{NeRF},
\end{aligned}
\end{equation}
where $\lambda_1$ and $\lambda_2$ are loss weights.

To bridge the gap between latent space and image space~\cite{muller2023diffrf}, we formulate $\mathcal{L}_\text{diff-img}$ with approximation $\bm{\tilde{z}}_{0}(\bm{\epsilon},\Phi):= \bm{z}_{0} + \frac{\sqrt{1-\bar{\alpha}_{t}}}{\sqrt{\bar{\alpha}_{t}}} (\bm{\epsilon} - \bm{\epsilon}_{\Phi}(\bm{z}_{t}, t, \bm{c}))$,
\begin{equation}
\label{eqn:diff_img_loss}
    \begin{aligned}
        \mathcal{L}_\text{diff-img}=w_t\mathbb{E}_{\bm{z}_{t}, \bm{\epsilon}, t, \bm{c}}\Arrowvert\mathcal{D}(\bm{\tilde{z}}_{0}(\bm{\epsilon},\Phi)) - \bm{I}^\text{tgt} \Arrowvert^{2},
    \end{aligned}
\end{equation}
where $\bar{\alpha}_{t}=\prod_{i=1}^{t}\alpha_i$, $\bm{I}^\text{tgt}$ is the image at the target view with the target human pose, $\mathcal{D}$ denotes the decoder in a VAE~\cite{kingma2013auto, van2017neural} model.
Since the approximation holds reliably only for steps $t$ close to zero, we introduce a weight $w_t$ that progressively decays as the step value increases. 
In addition to MSE loss, we apply structural similarity index (SSIM)~\cite{wang2004image} and Learned Perceptual Image Patch Similarity (LPIPS)~\cite{zhang2018unreasonable} as additional image-level loss terms.
For features from Human NeRF, we follow~\cite{chan2022efficient} to add a term $\mathcal{L}_\text{NeRF}$, which regularizes the first three channels by computing MSE, SSIM, and LPIPS with the target image $\bm{I}^\text{tgt}$ and a Mask Loss with the target human mask.

\section{Experiments}

\subsection{Experimental Setup}

\noindent \textbf{Datasets.} 
We evaluate the performance of our \nickname{} on four human modelling datasets, \ie, RenderPeople~\cite{renderpeople} (multi-view image), DNA-Rendering~\cite{cheng2023dna} (multi-view video), THuman2.1~\cite{tao2021function4d} (3D scan), and TikTok~\cite{jafarian2021learning} (monocular video).
For RenderPeople, we randomly sample 450 subjects as the training set and 30 subjects for testing.
For each subject, we use all frames from the training data for training, 6 frames with 4 camera viewpoints for novel view synthesis, and all frames with a front camera view for novel pose synthesis. 
For the DNA-Rendering dataset, we use 416 sequences from Part 1 and 2 for training and 10 sequences for evaluation.
For each subject, we use all frames from the training data for training, 48 frames with 4 camera viewpoints for novel view synthesis, and all frames with a front camera view for novel pose synthesis. 
The foreground masks, camera, and SMPL parameters from these two datasets are used for evaluation purposes.
For the THuman2.1 dataset, we randomly select 2345 3D scans for training and 100 3D scans for evaluation. 
We render 24 multi-view images at a resolution of 512x512 for each scan, and evaluate the performance on all rendered views of the test set.
For the TikTok dataset, we follow previous work~\cite{wang2023disco} to split the training and test sets. 
We estimate SMPL and camera from a 2D image using 4D-Humans~\cite{goel2023humans}, and segmentation masks from SAM~\cite{kirillov2023segment}.
For THuman2.1 and TikTok datasets, we perform the training on both datasets.

\noindent \textbf{Comparison Methods.}
We compare our \nickname{} with two categories of state-of-the-art single-view-based animatable human modelling methods, \ie, a generalizable Human NeRF method, SHERF~\cite{hu2023sherf}, and diffusion-based 2D character animation methods, AnimateAnyone~\cite{hu2024animate}, MagicAnimate~\cite{xu2024magicanimate} and Champ~\cite{zhu2024champ}.
We evaluate the performance of MagicAnimate and Champ by using their released checkpoints. 
For fair comparisons, we also fine-tune SHERF and Champ by using their official codebase and AnimateAnyone with the open-source implementation from MooreThreads\footnote{https://github.com/MooreThreads/Moore-AnimateAnyone}.
More implementation details are shown in Appendix.

\begin{figure*}[t]
    \centering
    \includegraphics[width=17cm]{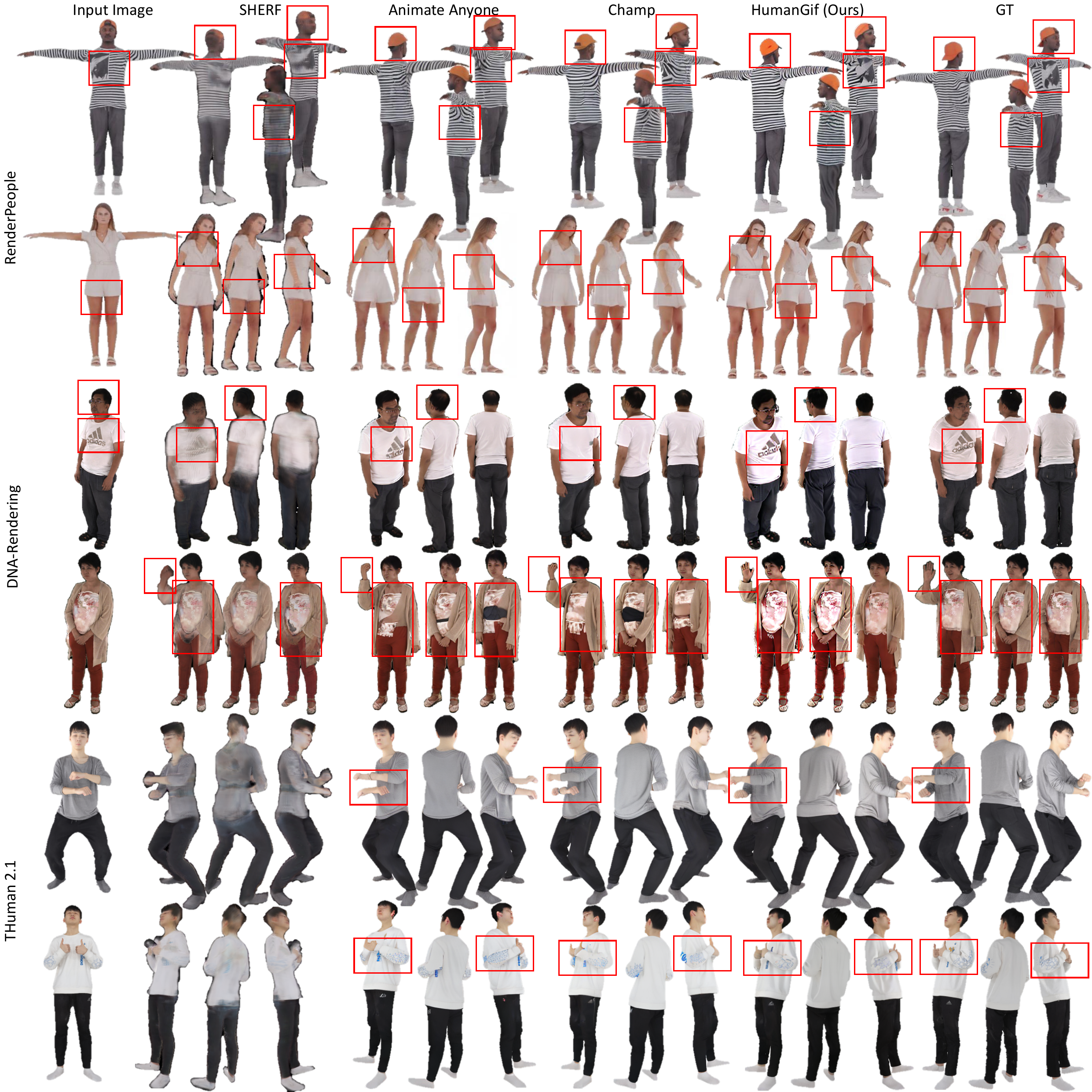}
    \setlength{\abovecaptionskip}{0cm}
    \caption{Qualitative results of novel view synthesis (1st, 3rd, 5th, and 6th row) and novel pose synthesis (2nd and 4th row) produced by SHERF, Animate Anyone, Champ, and our \nickname{} on RenderPeople, DNA-Rendering, and THuman2.1 datasets.} 
\label{fig: main_fig}
\vspace{-2mm}
\end{figure*}

\subsection{Quantitative Results}
As shown in Tab.~\ref{tab: main_result}, \nickname{} outperforms baseline methods in perceptual quality metrics (SSIM and LPIPS) across all datasets.
While SHERF, the state-of-the-art generalizable animatable Human NeRF method for single-view image input, achieves the highest PSNR scores in both novel view and pose synthesis tasks, it tends to produce blurred images, particularly for unseen views and poses. 
This is attributed to its limited capability to infer missing information from a single input image, which explains its subpar performance in perceptual quality metrics (SSIM and LPIPS).
To validate whether the original trained 2D SOTA character animation methods perform well in the novel-view and novel pose synthesis tasks, we evaluate the performance of MagicAnimate and Champ by using their released checkpoints. 
Although these methods achieve reasonable performance on novel pose synthesis, they fail to produce high-quality results on the novel-view synthesis task. 
To show fair comparisons, we fine-tune Animate Anyone and Champ on the same datasets as \nickname{}.
These two methods demonstrate reasonably good perceptual quality in unseen views and poses by leveraging the generative prior of Stable Diffusion.
However, they fail to capture detailed information from the input image.
Meanwhile, these methods fail to produce consistent novel view and pose results.
In contrast, our \nickname{} leverages proposed modules to effectively learn fine-grained details from the input image and produce consistent novel view and pose results, even in the single-image setting.
Beyond image quality metrics, we also evaluate video fidelity for animatable human videos generated by these methods. 
Our \nickname{} achieves the best FVD metric with a temporal-attention module, demonstrating superior temporal consistency. At the same time, SHERF performs reasonably well in video fidelity due to its effective LBS modeling.
We incorporate a user study in Fig~\ref{fig:user_study} to show human perceptual results.
To further show generalizability on in-the-wild images, we evaluate \nickname{} on in-the-wild human images by following the same data processing pipeline for TikTok.

\begin{figure}[h]
    \centering
    \includegraphics[width=9cm]{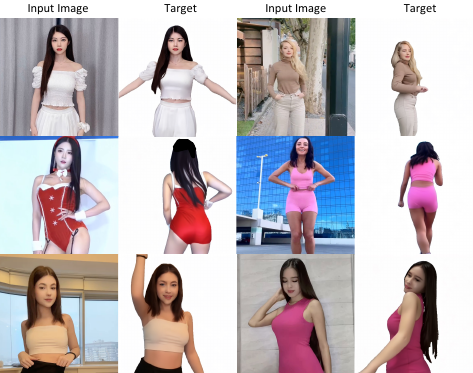}
    \setlength{\abovecaptionskip}{-2mm}
    \caption{Generalization performance of \nickname{} on in-the-wild images.} 
\label{fig: in_the_wild}
\vspace{-2mm}
\end{figure}

\begin{table*}[t]
    \centering
    \caption{
        Ablation study on DNA-Rendering. The left side shows examined combinations of components that are ablated.
    }
    \vspace{-6pt}
    \label{tab:ablation}
    \small{
    \addtolength{\tabcolsep}{-3.pt}
    \begin{tabular}{cccc|ccccccc}
        \toprule
        \multirow{2}*{\makecell{Generative \\ Prior}} & \multirow{2}*{\makecell{Human NeRF \\ Module}} & \multirow{2}*{\makecell{Image-Level \\ Loss}} & \multirow{2}*{\makecell{View/Motion \\ Attention}}& \multicolumn{3}{c}{Novel View} & \multicolumn{3}{c}{Novel Pose} \\
        \cmidrule{5-11}
        ~ & ~ & ~ & ~ &  PSNR$\uparrow$ & SSIM$\uparrow$ & LPIPS$\downarrow$ & PSNR$\uparrow$ & SSIM$\uparrow$ & LPIPS$\downarrow$ & FVD$\downarrow$\\
        \midrule
         & & & & 21.657 & 0.921 & 0.069 & 20.874 & 0.910 & 0.076 & 15.38 \\
        $\checkmark$ & & & & 23.684 & 0.860 & 0.056 & 23.080 & 0.870 & 0.061 & 17.96 \\
        $\checkmark$ & $\checkmark$ & & & \textbf{24.026} & 0.923 & 0.050 & \textbf{24.598} & 0.923 & 0.050 & 11.72 \\
        $\checkmark$ & $\checkmark$ & $\checkmark$ & & 22.731 & 0.927 & 0.050 & 23.376 & 0.930 & 0.049 & 11.59 \\
        $\checkmark$ & $\checkmark$ & $\checkmark$ & $\checkmark$ & 23.686 & \textbf{0.935} & \textbf{0.047} & 24.275 & \textbf{0.935} & \textbf{0.045} & \textbf{9.88}\\
        \bottomrule
    \end{tabular}}
\end{table*}

\begin{table}[h]
    \begin{center}
        \includegraphics[width=0.8\linewidth]{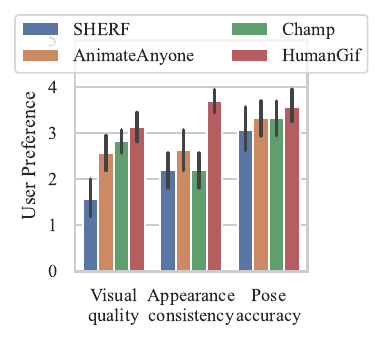}
    \end{center}
    \setlength{\abovecaptionskip}{0mm}
    \vspace{-14pt}
    \captionof{figure}{User preference scores.}
    \label{fig:user_study}
\end{table}

\subsection{Qualitative Results}
We show qualitative results of novel view synthesis (1st, 3rd, 5th, and 6th row) and novel pose synthesis (2nd and 4th row) of our \nickname{} and baseline methods in Fig.~\ref{fig: main_fig}. 
SHERF produces reasonable RGB renderings for the part visible from the input image, but it struggles to get realistic results for the part invisible from the input image.
For example, it produces blurry images in the back view when given a front-view image.
Thanks to the generative prior involved in our \nickname{}, we can produce realistic outputs (\eg, back view in the 3rd row of Fig.~\ref{tab: main_result}) in unseen views and poses.
2D character animation methods, Animate Anyone and Champ, produce realistic results, but they fail to recover fine-grained details (\eg, patterns on clothes and eyeglasses) from the input image.
By incorporating the Human NeRF module to enhance the information from the input 2D observation, our \nickname{} successfully recovers the detailed information from the input image.

\begin{figure}[t]
    \centering
    \includegraphics[width=9cm]{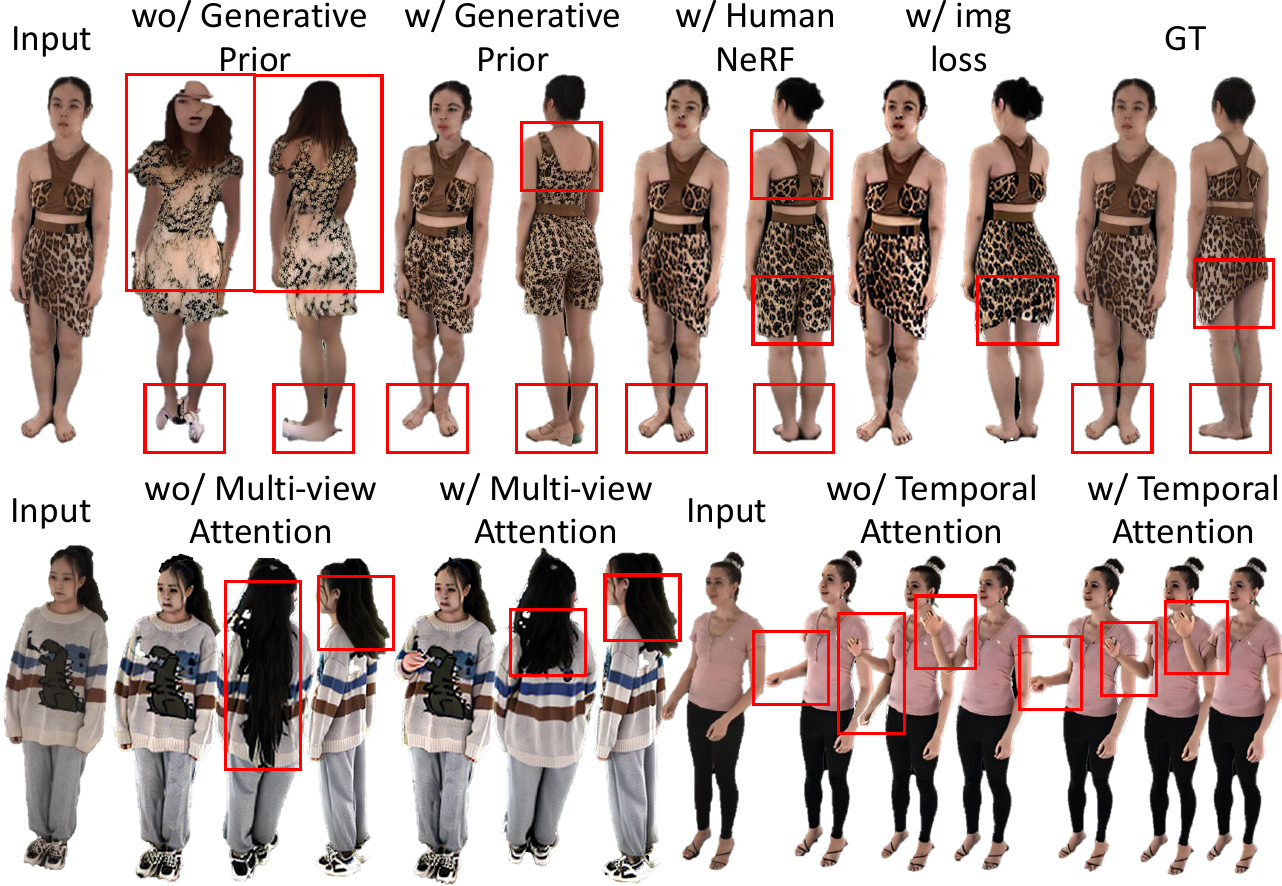}
    \setlength{\abovecaptionskip}{-2mm}
    \caption{Qualitative results of ablation study on DNA-Rendering dataset.} 
    \label{fig:ablation_fig}
\vspace{-4mm}
\end{figure}

\subsection{Ablation Study}
To validate the effectiveness of the proposed components, we subsequently integrate them and evaluate their performance on the DNA-Rendering dataset.
As shown in Tab.~\ref{tab:ablation} and Fig.~\ref{fig:ablation_fig}, training \nickname{} without inheriting the generative prior (pre-trained weights) from Stable Diffusion results in distorted human images with incorrect textures. 
By incorporating the generative prior, \nickname{} generates realistic human images while preserving the identity of the input. 
However, it still struggles to capture fine-grained details from the input image.
For instance, high-heeled shoes are mistakenly added to bare feet, and the garment style in the back view differs from that in the front view (see Fig.~\ref{fig:ablation_fig}). 
Incorporating the Human NeRF module enables \nickname{} to learn detailed information from the input image, effectively addressing these issues.
Additionally, introducing the image-level loss enhances the model's ability to produce consistent results. 
For example, pants are corrected to a skirt in the generated output.
Furthermore, the unified view/temporal attention mechanism ensures that the model learns view-consistent features, such as maintaining consistent hairstyles across views (as seen in Fig.~\ref{fig:ablation_fig}), and generates smooth novel pose sequences, and resolves issues related to inconsistent human poses.

\begin{figure}[h]
    \centering
    \includegraphics[width=0.9\linewidth]{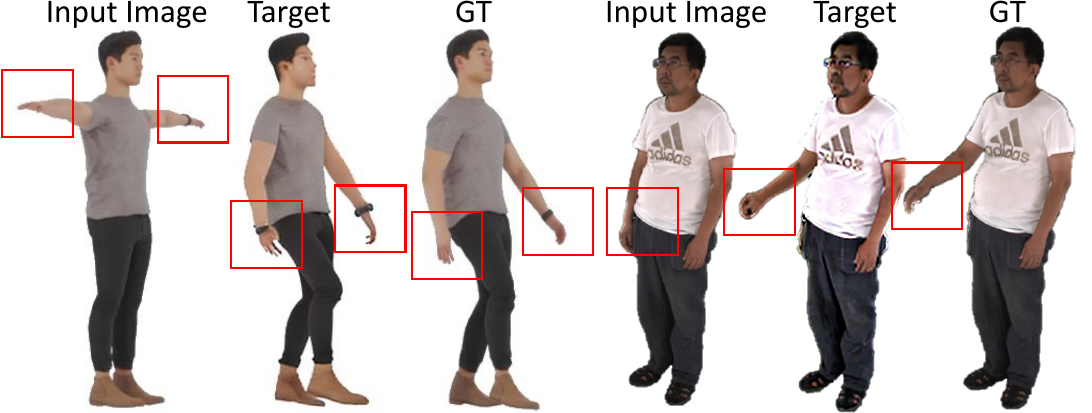}
    \setlength{\abovecaptionskip}{0cm}
    \caption{Failure cases on synthesis results.}
    \label{fig:failure_case}
    \vspace{-6pt}
\end{figure}

\section{Conclusion and Discussion}
To sum up, we propose \nickname{}, which reformulates the single-view-based 3D human novel view and pose synthesis as a single-view-conditioned human diffusion process, utilizing generative priors from foundational diffusion models to complement the missing information. 
To further produce perpetually realistic, view-consistent, and temporally coherent human avatars from a single image, we incorporate a spatially and temporally aligned conditional signal rendered from the Human NeRF module along with a Plücker ray representation, and a unified view/temporal attention layer into our \nickname{}.
Furthermore, we introduce an image-level loss during optimization to bridge the gap between latent and image spaces in diffusion models.
Experiments demonstrate that our \nickname{} achieves the best perceptual results. 

\noindent \textbf{Limitation and Future Work.}
1) While our proposed \nickname{} improves the synthesis performance, there still exists an inconsistency between the generated images and the input image. 
For example, as shown in Fig.~\ref{fig:failure_case}, it is still challenging to learn the correct geometry of fingers, and the generated images may contain additional accessories (\eg, watch).
How to align target images with the input image and target poses remains a future direction to be explored.
2) There is still room to improve the visual quality of generated results, especially for facial areas. 
We observe that VAE produces distortions for facial areas in some cases. 
Introducing VAE tailored for human images is a promising direction to improve the quality.
3) Our \nickname{} is trained on existing multi-view human datasets. 
Scaling our \nickname{} to larger in-the-wild human datasets remains a future direction to be explored.

{
    \small
    \bibliographystyle{ieeenat_fullname}
    \bibliography{main}
}

\clearpage
\appendix
\section{Implementation Details}
\noindent \textbf{Training Details.}
All our experiments are conducted on 4 NVIDIA H100 GPUs. 
The training phase is divided into three stages. In the first stage, we process all frames by resizing and cropping the images to be 512x512 resolution for the RenderPeople, THuman2.1, and TikTok datasets and 768x768 resolution for the DNA-Rendering dataset. 
Then we train a model to learn the mapping from a reference image and a target pose image to a target human image. 
During optimization, the Human NeRF module is jointly optimized with Denoising UNet, ReferenceNet, Pose Encoder modules.
This stage is trained for 120,000 iterations with a batch size of 8. 
In the second stage, we incorporate the view/temporal attention layer after the spatial-attention and cross-attention in the denoising UNet to learn multi-view consistency. 
We fine-tune the view/temporal attention layer while keeping the parameters of other modules fixed.
This stage is trained for 10,000 iterations with a batch size of 24 frames on each GPU. 
We inject a class embedding [0] for novel view synthesis and a class embedding [1] for novel pose synthesis.
We use the Adam~\cite{kingma2014adam} optimizer for all stages, set the initial learning rate as $1\times 10^{-5}$, and decay the learning rate with a linear scheduler.

\noindent \textbf{3D-aware features.}
We include two forms of 3D-aware features in our human NeRF module, \ie, point-level features and pixel-aligned features.
1) Point-level Features. 
We first extract per-point features by projecting the SMPL vertices onto the 2D feature map of the input image.
Next, we apply inverse LBS to transform the posed vertex features into the canonical space. 
These transformed features are then voxelized into sparse 3D volume tensors and further processed using sparse 3D convolutions~\cite{spconv2022}. 
From the encoded sparse 3D volume tensors, we extract point-level features $\bm{f}_\text{point}(\bm{x}^{c})$ for each point $\bm{x}^{c}$.
With the awareness of 3D Human structure, point-level features capture local texture details in the seen area and infer textural information for unseen areas through sparse convolution.
2) Pixel-aligned Features. 
Due to limited SMPL mesh and voxel resolution, point-level features suffer from significant information loss, especially in areas visible from the reference image.
To compensate for the information loss problem, we additionally extract pixel-aligned features by projecting 3D deformed points $\bm{x}^{c}$ into the input view.
Each deformed point $\bm{x}^{c}$ is transformed into the observation space as $\bm{x}^\text{ref} = \text{LBS}(\bm{x}^{c}; \bm{\theta}^\text{ref}, \bm{\beta}^\text{ref})$ using LBS. 
It is then projected onto the input view, allowing us to query the pixel-aligned features.
Leveraging the complementary strengths of point-level and pixel-aligned features, our Human NeRF module effectively captures fine-grained feature details in regions visible in the reference view while inferring features for regions occluded in the reference view.

\noindent \textbf{Evaluation Metrics.}
To quantitatively compare our \nickname{} with baseline methods, we evaluate the performance on three metrics, \ie, peak signal-to-noise ratio (PSNR)~\cite{sara2019image}, structural similarity index (SSIM)~\cite{wang2004image} and Learned Perceptual Image Patch Similarity (LPIPS)~\cite{zhang2018unreasonable}.
To further evaluate the video fidelity of animatable human videos produced from these methods, we follow~\cite{shao2024360} to report Fréchet Video Distance (FVD)~\cite{unterthiner2018towards}.
As the multi-view RenderPeople released from~\cite{hu2023sherf} does not contain animatable human videos, we omit the FVD metric in this dataset.

\end{document}